# TEXT RECOGNITION IN BOTH ANCIENT AND CARTOGRAPHIC DOCUMENTS

N. Zaghden [a,b], B. Khelifi [a,b], A.M Alimi [a], R. Mullot [b]

[a] REGIM: Research Group on Intelligent Machines, University of Sfax, ENIS, Department of Electrical Engineering, BP W - 3038, Rte Soukra km3.5, Sfax, Tunisia – (nizar.zaghden, khelifi_badreddine, adel.alimi)@ieee.org;
[b] L3i:Laboratoire Informatique Image Interaction, university of La Rochelle, Sciences Faculty, Av. Michel Crépeau La Rochelle Cedex 17042 France - remy.mullot@univ-lr.fr

**ABSTRACT:**

This paper deals with the recognition and matching of text in both cartographic maps and ancient documents. The purpose of this work is to find similar text regions based on statistical and global features. A phase of normalization is done first, in object to well categorize the same quantity of information. A phase of wordspotting is done next by combining local and global features. We make different experiments by combining the different techniques of extracting features in order to obtain better results in recognition phase. We applied fontspotting on both ancient documents and cartographic ones. We also applied the wordspotting in which we adopted a new technique which tries to compare the images of character and not the entire images words. We present the precision and recall values obtained with three methods for the new method of wordspotting applied on characters only.

## 1. WORD FEATURES EXTRACTION

### 1.1 Methodology

To characterize the textual contents of the images of ancient documents, we opted for describing the words or even the characters and not to work with a whole page of document. This is due to the fact of the nature of our application. In fact the images of documents which we have (Tunisian national library, base of madonne, British library) which have different properties from contemporary documents images. In more, the images than on whom we work did not contain the same noise and the words are also not normalized. Of this, the idea to work with imagettes words was born. In object to characterize textual content of our heterogeneous basis, we opted for beginning with the characterization of small entities in documents (characters, pseudowords…).

As first application, we chose to characterize blocks texts belonging to the same image after having cleaned the image by a gaussian filter. We followed the steps presented in the diagram presented in the figure 1.

### 1.2 Font Recognition

The optical font recognition known as OFR, is interested in the recognition) of the font of a writing. The optical recognition of characters is a domain which was already the object of multiple researches; however the main OCR systems (Optical Character Recognition) were applied to monofont texts. We were able to realize during our previous works so impressive results with wavelets and fractals for the characterization of printed Arabic fonts [Zaghden 2006] and we were even able to show the robustness of the new method that we adopted (CDB) to calculate the fractal dimensions of the images. This method allowed us to show its robustness in front of white gaussian noise especially if we compare it with the wavelets, which showed a net decrease in the rate of recognition of fonts.

We obtained in fact recognition rates lower than 30 % with wavelets on these degraded images.

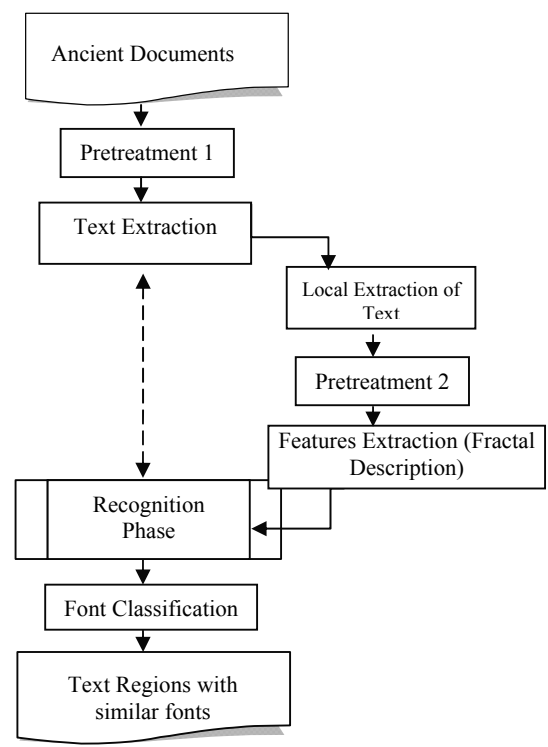

Figure 1: Organigram of the recognition font approach

In terms of recognition, we noticed a net performance of the new method which we adopted for the calculation of the fractal dimension (CDB). Indeed this method, allowed us to reach a recognition rate of 96.5 %. The obtained results prove that the methods which we adopted for the recognition of multifont texts can be applied to develop a strong character recognition system. Indeed the methods which we developed can be applied to a wider number of fonts and we prove in this paper that with this technique we can characterize different styles of texts contained in the same document.

The new method which we developed for the calculation of the fractal dimension (CDB) is inspired by the method of counting by Box [Zaghden 2006]. Among other by-products of this method, there is a method of counting of reticular cells [Gangepain 1986] and also the method of Differential Box Counting [Sarkar 1992, Sarkar 1994, Cho 99].

### 1.3 Characterization of textual documents by fractal dimensions

The robustness of the fractal dimensions was also illustrated in the characterization of different text styles of ancient documents where we notice the presence of three essential fonts (figure 3). We also demonstrated the possibility of this method to make the differentiation of scripts and fonts in the same image as it is shown in a cartographic map (figure2).

Certainly before making the characterization of the writing contained in the images we made first of all stages of pretreatment based on the works of [Khelifi 2007]:

1. Separation and reduction of colours
2. Elimination of the isolated pixels
3. Extraction of connected components
4. Segmentation of the textual constituents of the image

Figure 2: Extracted classes from cartographic image

Figure 3: Extracted classes from ancient document image

The inconvenience of this method is that we treat only the horizontal texts and we are now developing a system allowing us to correct the slope of texts to categorize in an effective way the various types of text which we can find in the various categories of ancient documents. This classification is of a big importance during categorization of the classes in several images.

## 2. TEXT MATCHING

### 2.1 Local Descriptors

After having applied global descriptors such as fractals to describe textual content of images, we chose to apply the matching of text images after having described them by local descriptors such as projection profiles, Euclidian Distance Map, XOR…

We can characterize various blocks text of the same size with the function XOR, in which we represent in the image result a white pixel only if it is a white pixel in one of the first two images and it is the opposite in the other image. So the function XOR works pixel by pixel. The calculation of the similarity between two images is done by applying this function between the new shape and the model in the basis. In the figure 18 we present the result of the function XOR on two occurrences of the same line.

So we can conclude that the function XOR can resolve the problem of recognition of the images words of ancient documents. Indeed we applied the Euclidian distance map algorithm to calculate the error between the images. Indeed every white pixel in the resultant image XOR error, so the function EDM applied to XOR allows us to obtain a vector measuring the error between two images.

This can simplify us the task especially if we choose to limit our application and to make a user interface in which, the user annotate for example words similar to the images words of our basis.

(a)

(b)

(c)

Figure 4: Application of XOR to two occurrences of the same line

### 2.2 Proposed Approach

The method that we just discussed allows us to make the wordspotting of the words images of ancient documents. The only problem is related in fact with the basis of images which we possess. George Washington's manuscripts possess several redundant words and many researchers worked on it [Manmatha 1996, Rath 2003], only the problem it is that we were not able to reach all the manuscripts. We chose to make the wordspotting on the images of the national library, Madonne and the base of Gutenberg and this is due to the fact that the words in our base are not too redundant. So to be able to give a significant result with many images in the base of test and learning, the samples which we treat, will be many. The method which we chose is not to work with a whole word but rather to work with the images of characters.

The documents on which we make our approach are only the ancient printed documents and essentially issued from the base of Madonne. Indeed, we make a first segmentation of the images based on the method developed in [Khelifi 2007] allowing to group the blocks of the images texts. Then we treat every block by choosing as criterion of separation between the characters to be segmented, the presence of white pixels. Certainly this approach allows to make the correspondence between characters on ancient printed documents but she will also allow to characterize pseudowords (Blobs) issued from the ancient handwritten documents in particular the Arabic characters printed or manuscripts owed to the fact of the nature of the Arabic writing, which is cursive by nature.

First of all we chose to work with the function XOR followed by the algorithm EDM to calculate the errors between the images requests and the images of the basis (figure 5); the images ( c ) and ( d ) present respectively the result and the complement to the result of the function XOR applied to the images ( a ) and ( b ).

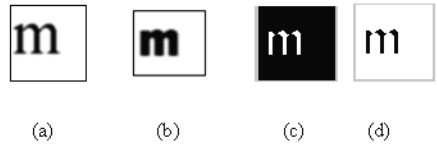

Figure 5: Application of XOR function to two occurrences of character "m"

Secondly we calculated the projection of profile of the images of characters.

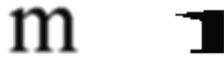

Figure 6: Vertical Profile of character "m"

The result of every character is compared in fact with the other characters of the document. So the advantage of our approach consists in the fact that we work with a not supervised classification, which applied in our domain of application will have certainly many advantages that the methods based on supervised classification.

One of the main advantages is that we do not know the number of classes or characters in advance whether it is in one image of ancient document or in all the basis. We became aware of this problem when we made Self Organizing Map's algorithms and K_means and we showed that the number of class can never be fixed from a document to the other especially if we work with images of heterogeneous documents belonging to various centuries and certainly having various characteristics.

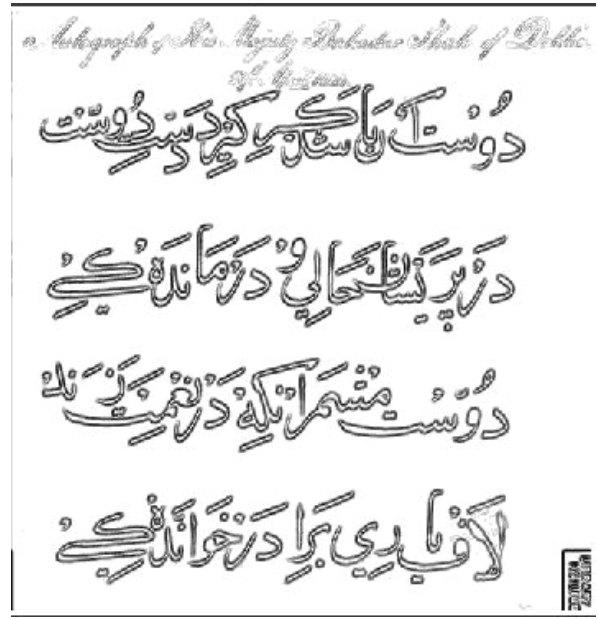

Figure 7: Application of The Self Organizing Map on an ancient document image

The images to which we applied our algorithms are the ones issued from the basis of madonne and of Gutenberg because the base of the national library contains much more noises and that the techniques that we developed are not rather effective on such degraded images. The images that we treat here contain the noise due to the weak resolution and also to the artefact resulting from the compression of collected images.

A phase of pretreatment is applied to the images (gaussian filter followed by the elimination of the isolated pixels) in the first place.

Our experiments touched only the images of Latin characters the number of which is 7280. Then we tested images of words of documents printed the number of which amounts to 100. These images tests repeat 135 times in the images considered in our base of learning (There are words which repeat more than only once).

We so calculated the terms precision and recall for two various methods which we adopted:

- XOR
- EDM (having applied XOR)
- Vertical Projection followed by the algorithm EDM

We present in the table below the results obtained with three described methods.

| Method | XOR | EDM | Vertical Projection |
|---|---|---|---|
| Precision | 63.14% | 78.43% | 41.67% |
| Recall | 55.34% | 79.32% | 34.46% |

Table 8: precision and recall rate of adopted methods

We can notice also that the best rates of recognition were obtained with the EDM method. Certainly this rate is the best in comparison with the methods which we tested but it will certainly be improved by adding the other criteria.

Among the works that we are making now, there is a correction and the improvement of the slope of the characters issued from ancient documents as well as the normalization and the centring of the characters. Other techniques also make our current domain of study such as the DTW (Dynamic Time Wrapping) function used for a long time in the analysis of the signals.

## 3. CONCLUSION

We tried in this paper to characterize the textual contents of the images of ancient documents. We tried several algorithms treating the pages of documents altogether such as the K-means algorithm and the algorithm "self organizing map", in order to categorize the various classes contained in our documents. We concluded that such methods cannot be taken as standard methods and effective capable of categorizing the textual entities of our documents. We also tried to apply global approaches for the measure of texture such as the fractal dimensions from the method CDB. We were able to categorize the various types of writings of our images. We also applied the wordspotting in which we adopted a new technique which tries to compare the images of character and not the word images or pseudowords. Some improvements are in the course of study to improve the rate of precision and recall.